\documentclass[conference]{IEEEtran}
\IEEEoverridecommandlockouts

\usepackage{graphicx} 
\usepackage{amsmath,amssymb,amsfonts}
\usepackage{blindtext}
\usepackage{wrapfig}
\usepackage{mathtools}
\usepackage{bbm}
\usepackage[colorlinks=true, linkcolor=blue, urlcolor=blue, citecolor=blue]{hyperref}
\usepackage{verbatim}
\usepackage{listings}
\usepackage{soul}
\usepackage[normalem]{ulem}

\usepackage{cite}
\usepackage{amsmath,amssymb,amsfonts}
\usepackage{algorithmic}
\usepackage{graphicx}
\usepackage{textcomp}
\usepackage[table]{xcolor}
\usepackage{booktabs}
\usepackage{multirow}
\usepackage{hyperref}
\usepackage[capitalise]{cleveref}
\usepackage{tabularx}
\usepackage{enumitem}
\usepackage{url}

\usepackage[utf8]{inputenc} 
\usepackage[T1]{fontenc}  
\usepackage{parskip}
\usepackage{subfigure}

\usepackage[acronyms,nonumberlist,nopostdot,nomain,nogroupskip,acronymlists={hidden}]{glossaries} 
\newglossary[algh]{hidden}{acrh}{acnh}{Hidden Acronyms}

\newacronym{tb3}{TB3}{Turtlebot3 Burger}
\newacronym{ckf}{CKF}{Cubature Kalman Filter}
\newacronym{mse}{MSE}{Mean Squared Error}
\newacronym{ros}{ROS}{Robot Operating System}
\newacronym{fps}{FPS}{Frames Per Second}
\newacronym{slam}{SLAM}{Simul-
taneous Localization and Mapping}

\allowdisplaybreaks

\def\BibTeX{{\rm B\kern-.05em{\sc i\kern-.025em b}\kern-.08em
    T\kern-.1667em\lower.7ex\hbox{E}\kern-.125emX}}
\begin{document}

\title{\huge Autonomous Robot for Disaster Mapping and Victim Localization\\
{\footnotesize Donatello Goes On A Rescue Mission}
\thanks{Identify applicable funding agency here. If none, delete this.}
}

\author{\IEEEauthorblockN{Michael Potter}
\IEEEauthorblockA{\textit{Cognitive Systems Lab} \\
\textit{Northeastern University}\\
Boston, USA \\
potter.mi@northeastern.edu} \\

\IEEEauthorblockN{Anuj Patel}
\IEEEauthorblockA{\textit{Masters in Robotics} \\
\textit{Northeastern University}\\
Boston, USA \\
patel.anuj2@northeastern.edu} \\
\and
\IEEEauthorblockN{Rahil Bhowal}
\IEEEauthorblockA{\textit{Masters in Robotics} \\
\textit{Northeastern University}\\
Boston, USA \\
bhowal.r@northeastern.edu} \\
\IEEEauthorblockN{Jingming Cheng}
\IEEEauthorblockA{\textit{Masters in Robotics} \\
\textit{Northeastern University}\\
Boston, USA \\
cheng.jingm@northeastern.edu} \\
\and
\IEEEauthorblockN{Richard Zhao}
\IEEEauthorblockA{\textit{Masters in Computer Science} \\ 
\textit{Northeastern University}\\
Boston, USA \\
zhao.richa@northeastern.edu} \\
}

\maketitle

\begin{abstract}

In response to the critical need for effective reconnaissance in disaster scenarios, this research article presents the design and implementation of a complete autonomous robot system using the \gls{tb3} with \gls{ros} Noetic. Upon deployment in closed, initially unknown environments, the system aims to generate a comprehensive map and identify any present 'victims' using AprilTags as stand-ins. We discuss our solution for search and rescue missions, while additionally exploring more advanced algorithms to  improve search and rescue functionalities. We introduce a \gls{ckf} to help reduce the \gls{mse} [m] for AprilTag localization and an information-theoretic exploration algorithm to expedite exploration in unknown environments. Just like turtles, our system takes it slow and steady, but when it's time to save the day, it moves at ninja-like speed! Despite Donatello's shell, he's no slowpoke - he zips through obstacles with the agility of a teenage mutant ninja turtle. So, hang on tight to your shells and get ready for a whirlwind of reconnaissance! \\ \\
Full pipeline code \href{https://github.com/rzhao5659/MRProject/tree/main}{https://github.com/rzhao5659/MRProject/tree/main} \\
Exploration code \href{https://github.com/rzhao5659/MRProject/tree/main}{https://github.com/rzhao5659/MRProject/tree/rz/exploration}
\end{abstract}


\glsresetall

\section{Problem Statement} 
\par In the wake of disasters, rapid and efficient reconnaissance is critical for successful rescue operations. Traditionally, human responders perform this task, but they often face considerable risks and limitations in accessibility. The development of autonomous robotic systems presents a promising alternative, capable of navigating and mapping disaster environments quickly and safely. However, existing autonomous systems primarily focus on mapping and fail to effectively integrate victim identification within their operational protocols.

\par This project we addresses the need for an autonomous robotic system that can both map unknown, closed environments and identify "victims(AprilTags)" in these settings. Our goal is to design and implement a robust system using the \gls{ros} Noetic, which can:
\begin{itemize}
    \item Generate a complete occupancy grid map of an unknown environment, providing essential data for navigation and strategic planning.
    \item Locate and accurately estimate the poses of "victims," represented by AprilTags, to simulate the identification and location of individuals in need of rescue.
\end{itemize}
\par The system must operate entirely autonomously, adapting to the dynamic and unpredictable nature of disaster environments. The key challenges include integrating advanced mapping techniques with efficient victim detection algorithms, ensuring the system’s reliability in diverse conditions, and managing computational constraints potentially imposed by the hardware used.

\section{Setup}  
\label{sec:setup}

The PC setup, Single Board Computer setup, and OpenCR setup were completed by adhering to the \gls{tb3} setup instructions provided on the Robotis website \cite{Robotis}. Subsequently, we progressed to more advanced software and hardware configurations after the initial setup.

\subsection{TurtleBot3 Hardware Setup} 
\par The \gls{tb3} is a widely used, compact, and customizable robotic platform suitable for a variety of robotic applications, particularly in educational and research environments. For our project, the \gls{tb3} platform was selected due to its modularity, ease of use, and compatibility with \gls{ros}. Below, we detail the hardware setup and configuration tailored for our autonomous reconnaissance system.
\subsubsection{Robot Model} We employed the \href{https://www.robotis.us/turtlebot-3-burger-us/}{\gls{tb3} model} for our project. This model is particularly advantageous for indoor environments due to its compact size and maneuverability. The key components of the \gls{tb3} include:
    \begin{itemize}
        \item Dynamixel X-series motors: These servos provide precise and reliable motion control.
        \item \href{https://emanual.robotis.com/docs/en/platform/turtlebot3/appendix_lds_02/}{360-degree LiDAR sensor(LDS-02)}: Essential for accurate mapping and obstacle avoidance.
        \item \href{https://emanual.robotis.com/docs/en/platform/turtlebot3/sbc_setup/}{Raspberry Pi}: Serves as the main processing unit, capable of handling both \gls{ros} operations and additional computation for sensor data processing.
    \end{itemize}
    
\subsubsection{Sensor Configuration}
To enhance the \gls{tb3}’s capabilities for detailed environmental mapping and victim identification, we configured the following sensor setup:
    \begin{itemize}
        \item \href{https://emanual.robotis.com/docs/en/platform/turtlebot3/appendix_lds_02/}{LiDAR}: Mounted at the top of the robot, providing a 360-degree field of view to detect obstacles and assist in generating a high-resolution occupancy grid map.
        \item \href{https://emanual.robotis.com/docs/en/platform/turtlebot3/appendix_raspi_cam/}{Camera}: Raspberry Pi Camera was attached to capture visual data, crucial for AprilTag detection. This camera is positioned to have an unobstructed view of the environment ahead of the robot.
    \end{itemize}
    
\subsubsection{Additional Hardware}
        \par Given the computational demands of our project, particularly for processing occupancy grids and detecting AprilTags, we supplemented the onboard Raspberry Pi with the following:
    \begin{itemize}
        \item External Battery Pack: To ensure extended operational periods without the need for recharging.
        \item Wi-Fi Adapter: To enhance the communication capabilities with the \gls{ros} network and enable off-board processing if required.
    \end{itemize}
    
\subsubsection{Calibration}
    \par Before deployment, the \gls{tb3} underwent a series of calibration procedures to ensure accuracy in its mapping and navigation capabilities:
    \begin{itemize}
        \item LiDAR Calibration: To verify the accuracy of distance measurements and angular resolution.
        \item Camera Calibration: Essential for precise AprilTag detection, involving adjusting the focus and verifying the camera’s field of view aligns with the expected operational environment. We followed the \gls{ros} tutorials for monocular camera calibration \cite{cameracalibration}.
    \end{itemize}

\subsection{TurtleBot3 Software Setup} 

Both TurtleBot and remote PC run on \gls{ros} Noetic.  For communication between these two, they need to be connected to the same network, and the remote PC need to run \textbf{roscore} to act as the \gls{ros} master.

On the TurtleBot, it's necessary to install the packages \verb|TurtleBot3_description|, \verb|TurtleBot3_bringup|, \verb|raspicam_node|, and \verb|apriltag_ros|. 
Our solution launches these files on the TurtleBot for bringup and AprilTags detection:
\begin{itemize}[nolistsep]
    \item \textit{TurtleBot3\_robot.launch} from TurtleBot3\_bringup.
    \item \textit{camerav2\_640\_624\_15fps.launch} from raspicam\_node.
    \item \textit{continuous\_detection.launch} from apriltag\_ros.
\end{itemize}  

The \textit{TurtleBot3\_robot.launch} starts up the RPLidar and the OpenCR board.  It publishes all sensor topics and subscribes to \verb|/cmd_vel| for controlling the wheel motors.  It also provides wheel odometry as a transform from frame \verb|odom| to frame \verb|base_footprint|, and as a message in \verb|/odom|.

The \textit{camerav2\_640\_624\_15fps.launch} starts up the Raspberry Pi Camera Module in 640x624 resolution with 15 \gls{fps} framerate. This publishes the raw image in \verb|/image_raw|, and the camera calibration matrix in \verb|/camera_info|. 

The \textit{continuous\_detection.launch} uses the raw camera output to detect and estimateAprilTagpose. It provides a transform from camera frame \verb|raspicam| to the tag frame \verb|Tag<id>|.  This information is also published in \verb|/tag_detections|.

On the remote PC, it's necessary to install the external packages \verb|TurtleBot3_description|, \verb|TurtleBot3_navigation|, \verb|TurtleBot3_simulations|, \verb|TurtleBot3_slam|, and \verb|m-explore|. Our solution launches these files on the remote PC to launch the autonomy stack:

\begin{itemize}[nolistsep]
    \item \textit{TurtleBot3\_slam.launch} from TurtleBot3\_slam.
    \item \textit{move\_base.launch} from TurtleBot3\_navigation.
    \item \textit{explore\_lite.launch} from m-explore.
    \item \textit{ckf3D} node.
    \item \textit{search and rescue} node.
\end{itemize}  

The \textit{TurtleBot3\_slam.launch} launches gmapping to perform \gls{slam} using LIDAR scans, and a robot state publisher node to publish all sensor transforms relative to the robot frame \verb|base_link|, which are specified in the URDF file.  The gmapping node provides an occupancy grid map in \verb|/map| as well as a transform from frame \verb|map| to frame \verb|odom|.  This completes the transform tree, and allows us to obtain the robot's current pose relative to the map frame. 

\begin{figure*}[ht!]
    \centering
    \subfigure[Turtlebot3 house]{\includegraphics[width=0.32\textwidth]{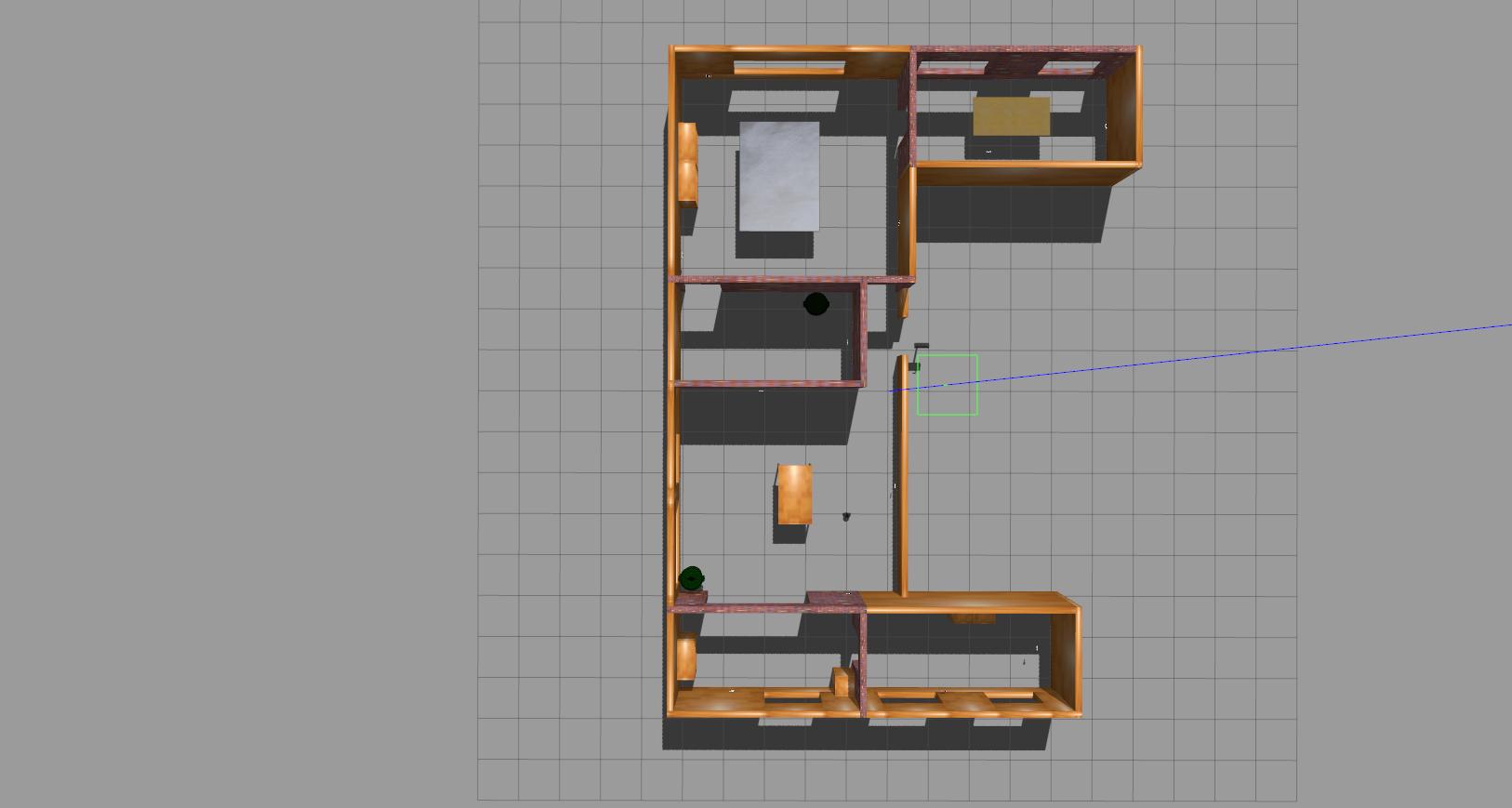}} 
    \subfigure[Turtlebot3 world]{\includegraphics[width=0.32\textwidth]{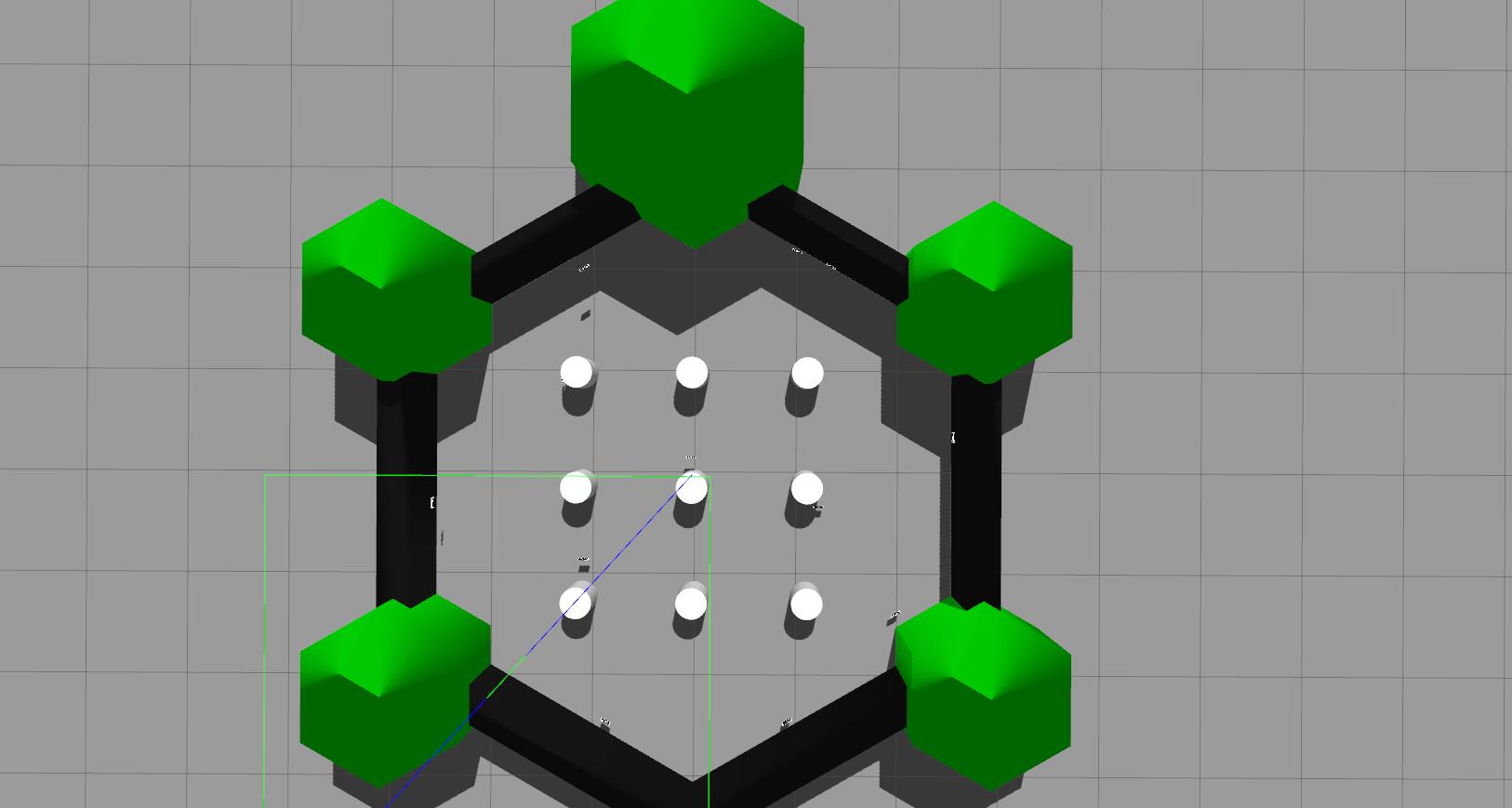}} 
    \subfigure[Maze]{\includegraphics[width=0.28\textwidth]{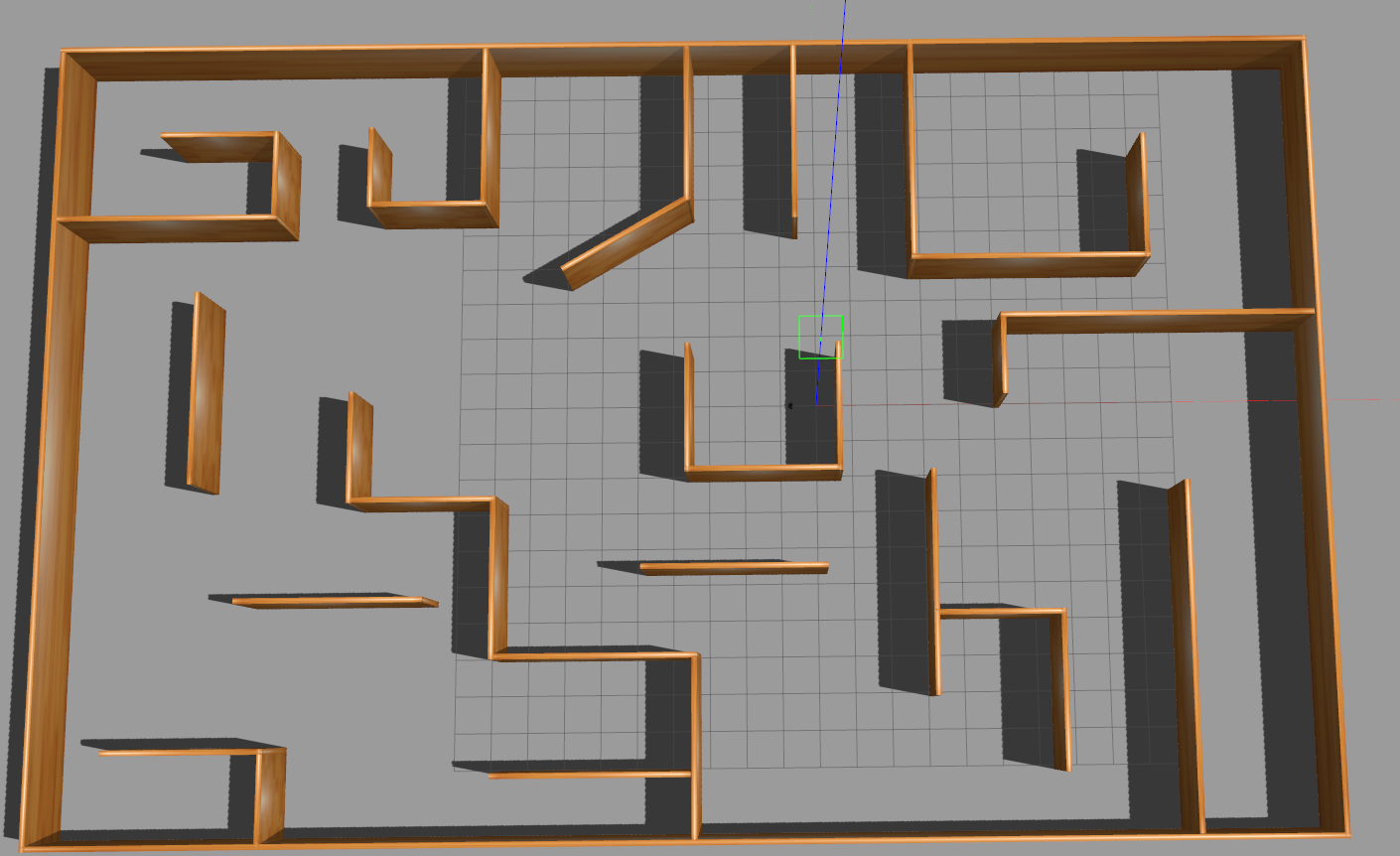}}
    \caption{Gazebo simulation arenas. (a) is 15.0m$\times$10.0m (b) is 3.0m$\times$3.0m (c) is 34.5m$\times$21.75m.}
    \label{fig:gazebo_arenas}
\end{figure*}

The \textit{move\_base.launch} launches move base node, which performs planning and control.  It subscribes to an action topic \verb|move_base/goal| that specifies the goal, plans a path using the default global and local planners, and publishes back the command velocity to \verb|/cmd_vel| topic, which the \gls{tb3} listens to.  The move base utilizes the occupancy grid map \verb|/map| obtained from SLAM and inflates the obstacles seen to produce a costmap, where the planning is performed.   

The \textit{explore\_lite.launch} runs a greedy frontier-based algorithm to compute the next goal for exploration. It receives the occupancy grid map either from gmapping or the costmap from move\_base, and publishes the exploration goal in \verb|move_base/goal|. 

These three nodes complete the essential autonomy stack and enables our robot to explore and map an environment autonomously, thus completing the first task of the problem. The remaining two nodes are used for the second task of the problem, which is identification and pose estimation of the "victims" (represented as AprilTags). 

The \textit{ckf3D} node launches a \gls{ckf} \cite{sarkka2023bayesian} that continuously receives the last pose estimates from theAprilTagdetector in \verb|/tag_detections| and outputs a filtered pose estimate for each identified tag as new transforms.  This improves the accuracy of pose estimate for the second task.

The \textit{search\_and\_rescue} node is the main algorithm that runs after the exploration node finishes its task, indicated through a \verb|/exploration_stop_flag| topic. This computes a series of goals that covers uniformly the explored map and order them in a queue such that the robot would move in a snake-like (zig zag) search pattern. These will then be sent to move base through the action topic \verb|move_base/goal|. In addition, this node directly commands the TurtleBot to spin after reaching each goal.  This node will allow the robot to identify any missing AprilTag during the exploration, and improve the pose estimation of AprilTags by increasing the number of pose measurements for the \textit{ckf3D} node. Once the queue is empty and all the AprilTags are processed, the \textit{search\_and\_rescue} node publishes the \verb|/finish_search| topic, to which the \textit{ckf3D} node subscribes. Subsequently, the \textit{ckf3D} node saves all the AprilTag positions in two JSON files: one containing the \gls{ckf} estimates and the other containing the \verb|apriltag_ros| package estimates.

 For simulating the \gls{tb3}, we launch \textit{continuous\_detection.launch}, the autonomy stack discussed, and one of the files inside the \verb|TurtleBot3_gazebo| package on the remote PC, such as TurtleBot3\_houseAT.launch.  Each launch file simulates a different gazebo environment.  Launch files ending with AT.launch are gazebo environments that contain AprilTags, which allows a complete simulation of our solution.

\section{Environment} 
\subsection{Gazebo World}
For the experimental setup, we created two Gazebo\cite{gazebo} environments to simulate a search and rescue mission with AprilTags serving as victims. A total of 12 AprilTags were set up, each measuring around 100mm x 100mm. Each tag has a unique ID between 0-11 and belongs to the family of \textit{April tag 36h\_11}\cite{apriltag_ros}. The reference worlds were turtlebot3\_house as shown in \cref{fig:gazebo_arenas}(a) and turtlebot3\_world  as shown in \cref{fig:gazebo_arenas}(b). A third Gazebo environment (\cref{fig:gazebo_arenas}(c)) was created to compare \textit{explore\_lite} package and our exploration implementation.

\subsection{Real World}
\begin{figure}[h!]
    \centering
    \includegraphics[width=4.5cm]{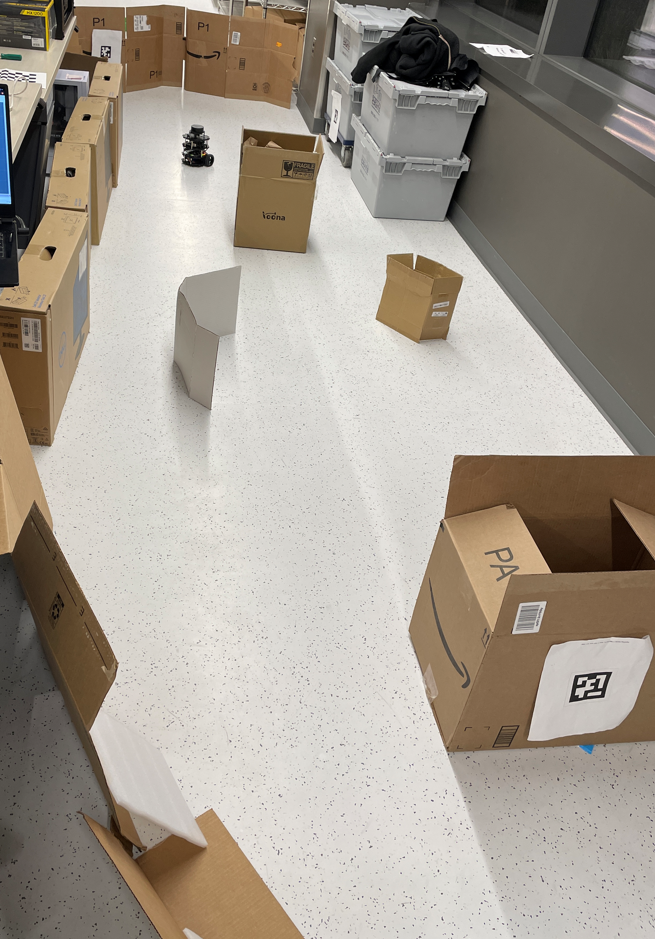}
    \caption{Example real-world arena setup. Please note that the final setup is larger and includes more AprilTags, as depicted in the demo videos.}
    \label{fig:realworld}
\end{figure}

For the experimental setup in the real world, we attempted to set up the arena using cardboard and placed 8 AprilTags belonging to the \textit{April tag 36h\_11} family, each with a unique ID between 0-7. The maze setup is shown in \cref{fig:realworld}.

\section{Methodology}
\begin{figure*}[ht!]
    \centering
    \includegraphics[width=17cm]{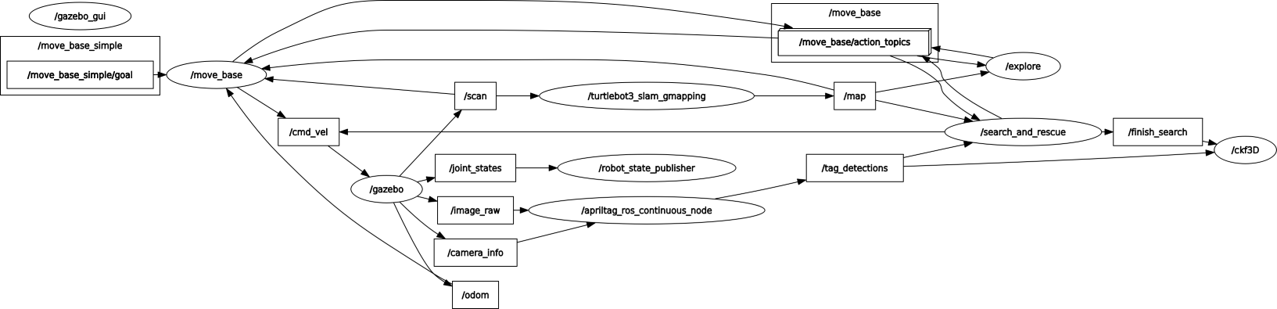}
    \caption{Nodes and topics in our solution}
    \label{fig:rqtgraph}
\end{figure*}
\subsection{Proposed Solution}
As outlined in \cref{sec:setup}, our initial setup utilized existing packages such as \verb|explore_lite|\cite{explore_lite} for frontier exploration, the \verb|move_base|\cite{move_base} action server for navigation, \verb|apriltag_ros|\cite{apriltag_ros} for AprilTag detection, and \verb|gmapping|\cite{gmapping} for \gls{slam}. Additionally as described in \cref{sec:setup}, custom nodes such as \textit{ckf3D} were employed for recursive Bayesian estimation of AprilTag position, and \textit{search\_and\_rescue} ensured the entire map was searched again to detect all the AprilTags.

To navigate the \gls{tb3} through the environment while effectively mapping and detecting AprilTags, we integrated the \gls{ros} package \verb|explore_lite|. Utilizing frontier-based exploration, this package aided in generating a comprehensive map. However, a key challenge arose due to potential oversight of AprilTags during exploration, attributed to the limited field-of-view of the Raspberry Pi Camera. Our attempt to address this involved adjusting parameters within \textit{gmapping}, specifically the minimum and maximum scan distances, under the assumption that these adjustments would enable simultaneous exploration for frontiers and AprilTag detection. However, this approach proved inefficient in achieving the desired outcome.

With \verb|explore_lite| operating as a greedy frontier-based exploration package, prioritization of large frontiers occurs. The influence of frontier size on the frontier cost is determined by parameters such as gain scale and potential scale. By elevating the potential scale to a value exceeding its default of 3, we aimed to prevent the TurtleBot from diverting from its path to pursue distant, larger frontiers, thereby promoting completion of current paths before exploration of new ones.

Furthermore, setting a minimum frontier size, particularly when utilizing \verb|explore_lite| alongside a \gls{slam}-generated map, accounts for potential \gls{slam} mapping oversights, especially in obstacle detection. This parameter specification prevents pursuit of frontiers that may be inaccessible or impractical to explore.

In our solution, we modified only the minimum frontier size while keeping other parameters default. These baseline solutions were integrated with our custom \gls{ros} nodes \textit{ckf3D} and \textit{search\_and\_rescue}. The \gls{ros} communication network, illustrated in \cref{fig:rqtgraph}, depicts interactions among all the nodes and topics pertinent to the autonomous search and rescue stack. This network facilitates seamless communication among software components within the \gls{ros} framework \cite{ros}.

\subsection{Exploration} 

The \verb|explore_lite| package is used for real world demo. It's a simple frontier-based exploration algorithm that picks the lowest score frontier centroid as the exploration goal.  A frontier's cost is computed as a weighted difference between the robot's distance to its closest frontier cell and the frontier's size (number of frontier cells).

It is able to achieve exploration, but has some limitations:
\begin{itemize}[nolistsep]
    \item The chosen exploration goal's orientation is left as zero, which is not optimal for discovering unknown areas, especially if the sensor has limited FOV sensor, such as camera. 
    \item The chosen exploration goal might be unsafe or untraversable: A frontier's centroid may be located in unknown or occupied space. This can cause \verb|explore_lite| to be stuck until a lack of progress timeout is reached, then the node blacklists that goal.
    \item The frontier detection is not efficient:  Every iteration starts anew and performs an entire map search for frontier cells, starting from the closest free cell. This isn't noticeable in small to medium sized maps as the planner frequency is 0.33 Hz, but it's expected to be slow for larger maps. 
\end{itemize}

We implemented a more efficient and effective exploration algorithm that combines ideas from both frontiers and next best view exploration methods. 

The exploration algorithm consist of two parts: a) frontier detection and b) computation of exploration goal. The frontier detection is responsible for detecting new frontier cells and forming them into frontiers. This runs continuously in the background. The computation of exploration goal is responsible to return an exploration goal based on the detected frontiers. This runs whenever the associated service is requested. 

\begin{figure}[h!]
    \centering
    \includegraphics[width=0.5\linewidth]{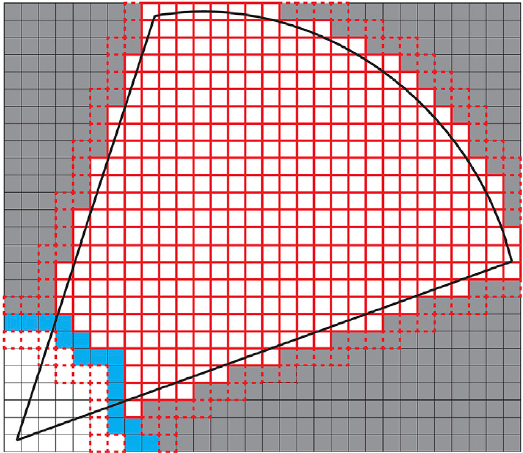}
    \caption{Expanding Wavefront Frontier Detection. Blue cells are frontier cells from previous timestep. Red cells are new, unexplored free cells that are traversed and evaluated as potential frontier cells. Figure from \cite{approaches-frontiers}.}
    \label{fig:ewfd}
\end{figure}

\textbf{Frontier Detection:} The frontier detection implements the algorithm Expanding Wavefront Frontier Detection (pseudo-code in \cite{ewfd}). It is a BFS algorithm that traverses free cells on the map to detect new frontier cells, as shown in \cref{fig:ewfd}.  Each free cell inserted into the queue is marked as visited and evaluated to ascertain its status as a frontier cell.  If it's a free cell and it has at least one unknown neighbor, then it's considered to be a frontier cell. Every detected frontier cell is then stored inside a data structure that only contains frontier cells. 

The efficiency of this algorithm comes from these two ideas:
\begin{itemize}[nolistsep]
    \item The visited state is incorporated as an additional layer to the occupancy grid map, so it persists across iterations. Hence, at each iteration, the algorithm only need to search the new, unexplored free cells that are uncovered due to new observations.  
    \item The BFS doesn't naively start searching from robot's current position for unexplored free cells. It starts from frontier cells in the previous timestep $F_{t-1}$ within the observation area $A_t$.  It exploits the fact that new frontier cells can only appear within the region the robot observes and beyond the past frontier cells.
\end{itemize}

The algorithm requires constant insertion and removal of frontier cells, as well as a quick range query of frontiers cells within the observation area. For this reason, we used a RTree data structure to store frontier cells, which is a spatial indexing structure that supports quick range query, similar to KD-tree, but it additionally supports dynamic insertion and removal with self-balancing. We use a bounding box of width and height of the \verb|OccupancyGridUpdate| message from the global costmap as the observation area.

Once the new set of frontier cells $F_t$ has been determined, we perform another search to form frontiers from connected frontier cells.  Our implementation of frontiers stores centroid's position and size (number of frontier cells). 

\textbf{Computation of Exploration Goal:}  As discussed, choosing a position among frontier centroids can have several limitations.  For this reason, we used ideas from next best view planners to sample around free space for an optimal exploration goal. 

\begin{figure}[h!]
    \centering
    \includegraphics[width=0.274\linewidth]{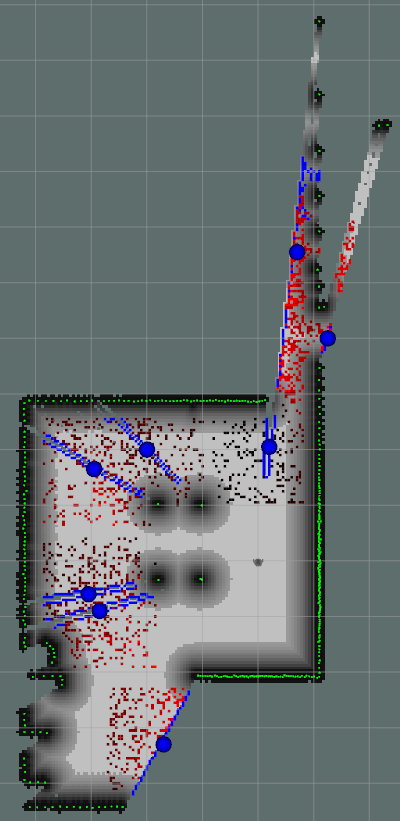}
    \includegraphics[width=0.5\linewidth]{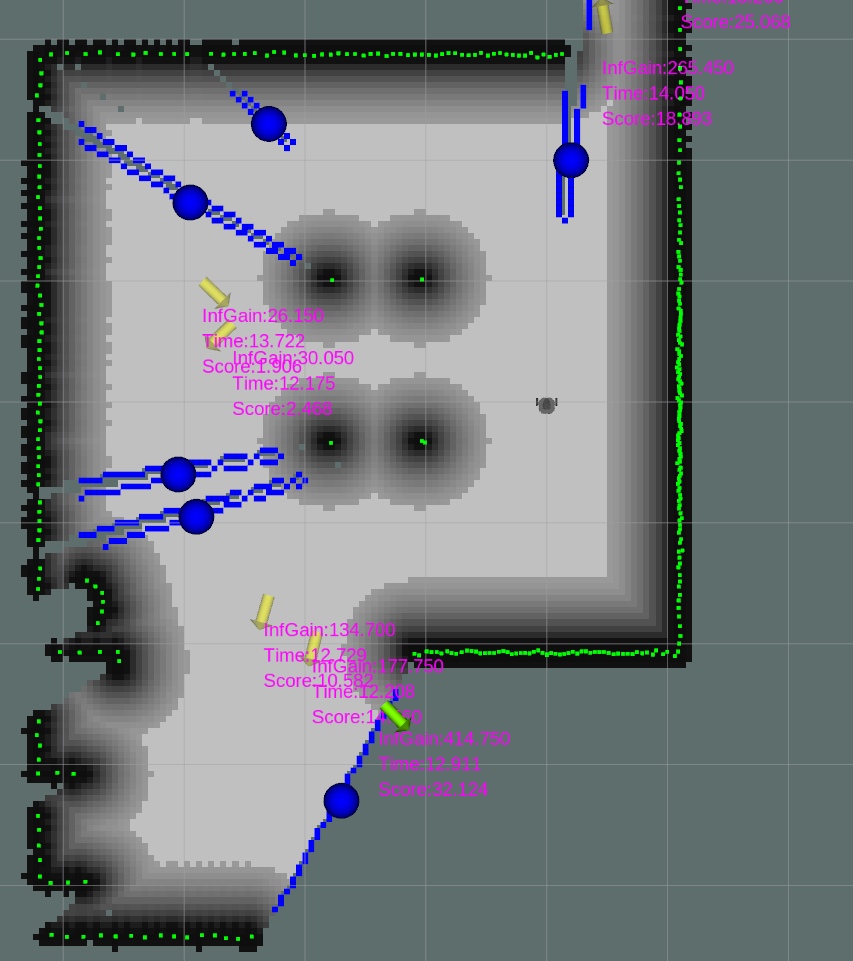}
    \caption{Computation of exploration goal. a) Sampling process b) Potential gain.  The yellow arrows indicate best pose near each frontier, and the green arrow indicate the best pose overall.}
    \label{fig:goal_exploration}
\end{figure}

The algorithm for choosing an exploration goal is as follows:
\begin{enumerate}[nolistsep]
    \item Sample $M$ points around each frontier's centroid in the free space (\cref{fig:goal_exploration}).
    \item Choose an optimal orientation for each sampled point that maximizes the information gain.  
    \item Compute the potential gain for each pose. 
    \item Choose the pose with highest potential gain as the exploration goal (\cref{fig:goal_exploration}).
\end{enumerate}

We compute the total unknown area that the robot would discover as the information gain, similar to \cite{nbv2002}. The information gain and the orientation optimization are implemented according to the next algorithm \cite{fast-nbv}:
\begin{enumerate}[nolistsep]
    \item Discretize $360^\circ$ into rays.
    \item For each ray $i$, count the number of unknowns cells $N_i$. The information gain for this ray is computed as the length of unknown cells: $I_i = N_i \cdot \text{map resolution}$
    \item For each orientation $\theta$, the information gain is computed by summing up each ray's information gain in the sensor's field of view:
    \begin{align}
    \text{I}_{\theta} = \sum_{i \in \text{sensor fov}}{I_i}        
    \end{align}
    \item Choose orientation $\theta^*$ that maximizes information gain.  
\end{enumerate}

The potential gain $G(X)$ for each pose $X = \{\textbf{p}, \theta\}$ is computed using information gain and time cost.  Let the robot's current pose be denoted as $X_r = \{\boldsymbol{p_r}, \theta_r\}$ and its maximum velocities be $\{\omega_{max}, v_{max}\}$, the potential cost is given by: 

\begin{figure*}
    \centering
    \includegraphics[width=14cm]{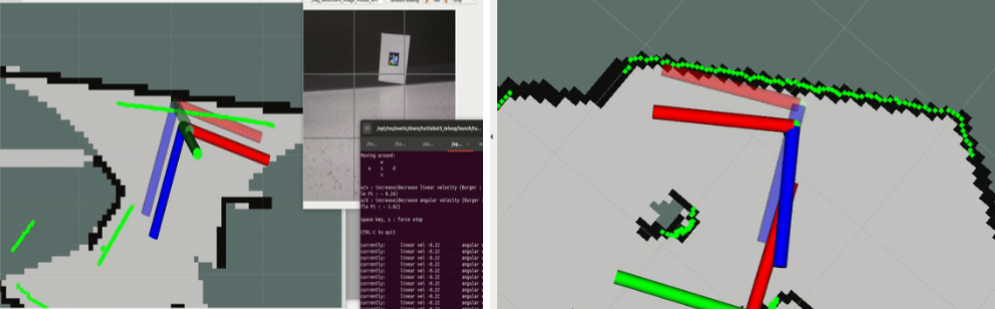}
    \caption{The lightly shaded axes represent the correct \gls{ckf} estimator for the AprilTag position, while the darker shaded axes indicate the incorrect apriltag\_ros estimator, which should align flush against the wall in RVIZ. The left figure is the real world arena, whereas the right figure is the Gazebo \gls{tb3} World arena (using the virtual camera software plugin)}
    \label{fig:estimator_bias}
\end{figure*}

\begin{figure}[h!]
    \centering
    \includegraphics[width=7cm]{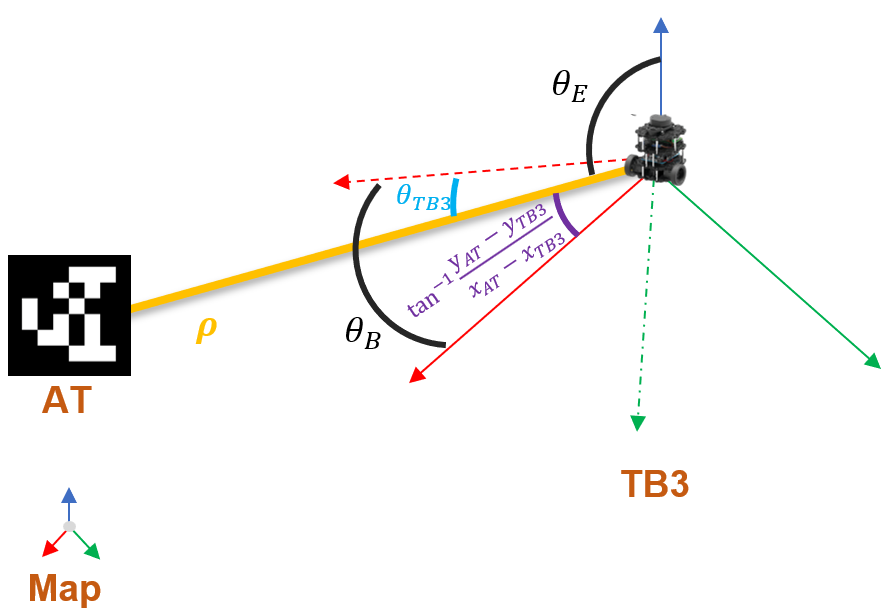}
    \caption{The geometry of the measurement model when using the World Coordinate Frame and the \gls{tb3} Coordinate Frame}
    \label{fig:geometry}
\end{figure}

\begin{align}
    G(X) &= \frac{I(X)}{T(X)} \\
    T(X) &= \text{max}\left(\frac{\theta - \theta_{r}}{\omega_{max}} , \frac{\|\boldsymbol{p} - \boldsymbol{p_r}\|}{v_{max}} \right)
\end{align}

 The time cost $T(X)$ is a simple estimate that assumes there is a straight, collision-free path toward the given pose, and the robot is moving with maximum velocities.  The information gain $I(X)$ is computed as mentioned.  When scoring sampled points, we use their optimal orientation. 

\subsection{AprilTag Pose Estimation} 
Qualitatively, we observed that the \verb|apriltag_ros| package displayed estimator bias in estimating the AprilTag position. As the distance between the \gls{tb3} and an AprilTag increased, the estimated AprilTag position drifted farther from the true position in the real world, as observed in RVIZ. To address this issue and other concerns outlined in \cref{sec:results}, we implemented a recursive Bayesian estimator, specifically the \gls{ckf} \cite{sarkka2023bayesian}, for the position estimate of each AprilTag in the World frame.

We utilize the \verb|apriltag_ros| package's relative transformation between the Raspberry Pi camera (raspicam) and the AprilTag to construct the measurement model for the 3D AprilTag position, considering range ($\rho$), bearing ($\theta_B$), and elevation ($\theta_E$) (see \cref{fig:geometry} for more details). Leveraging the \gls{ros} \verb|tf| package, we can transform the AprilTag and \gls{tb3} positions into the World coordinate frame. Since the AprilTags remain stationary, we assume the transition function for the AprilTag is an identity with zero noise. Thus, we may describe the state-space conditional probabilities required for the \gls{ckf} AprilTag3 position estimator as:
\begin{align}
    p\left(X_{AT}^{(k)} | X_{AT}^{(k-1)}\right) &= X_{AT}^{(k-1)} + \mathcal{N}(0,0) \\
    p\left(z^{(k)} | X_{AT}^{(k)} ; X_{TB3}^{(k)}\right) &=  h\left(X_{AT}^{(k)};X_{TB3}^{(k)}\right)  + \mathcal{N}(0,R^{(k)})
\end{align}
where we define measurement mean $h$ and measurement noise covariance $R^{(k)}$ as
\begin{align}
    h(\chi_AT;\chi_{TB3}) &= \begin{bmatrix}
        \theta_B \\ \theta_E \\ \rho
    \end{bmatrix} = \begin{bmatrix} \tan^{-1} \left( \frac{y_{AT} - y_{TB3}}{x_{AT} - x_{TB3}}\right) - \theta_{TB3} \\ \cos^{-1} \frac{\Delta Z}{\rho} \\
    \sqrt{\Delta X^2 + \Delta Y^2 + \Delta Z ^2}     \end{bmatrix}
\\
    R^{(k)} &= \begin{bmatrix}
       0.05 & 0 & 0 \\
       0 & \frac{\pi}{20} & 0 \\
       0 & 0 & \frac{\pi}{20}
    \end{bmatrix} \left( \rho^{(k)} \right)^4
\end{align}

In the \gls{tb3} coordinate frame, $\theta_{TB3}$ represents the positive angle from the x-axis to the AprilTag, while $\Delta X$, $\Delta Y$, and $\Delta Z$ denote the position of the AprilTag. The geometry of the \gls{tb3} and the AprilTag to describe the measurement model is shown in \cref{fig:geometry}

As the distance between the \gls{tb3} and the AprilTag increases, the measurement model covariance matrix also increases, reducing the impact of erroneous measurements on the final position estimate. Additionally, employing repeated measurements helps resolve occlusion issues, as earlier (but not necessarily the last) measurements may have zero occlusion.

\subsection{Search and Rescue} 

To ensure all AprilTags are detected in the environment, our custom node \textit{search\_and\_rescue} waits for the \verb|explore_lite| package to finish exploring the environment. Once the \verb|exploration_flag| is received, indicating the completion of the exploration phase and the start of the search and rescue operation, the approach utilizes a grid-based decomposition of the environment represented by the occupancy grid map.

The path planning process includes the following key steps:

\begin{enumerate}[nolistsep]
\item Inflated Occupancy Grid: The occupancy grid map undergoes inflation to accommodate the robot's physical dimensions, ensuring obstacles are buffered by a safety margin during path planning.
Free Space Identification: The inflated map is analyzed to identify unblocked areas, which constitute the free space accessible to the robot.
\item Grid Division and Sampling: The free space is divided into a grid structure, with the number of grid cells determined based on the estimated free space area $|grid\;cells| = \sqrt[3]{A_{free}}$. Each cell is iterated upon, and sampling points are generated within its boundaries.
\item Valid Point Selection: Sampled points are evaluated against the original occupancy grid map to confirm they reside in free space. If a sampled point falls within an obstacle, the algorithm attempts to find a valid point within the same cell through additional random sampling.
\item World Coordinates and Visualization: Valid points, confirmed to be free of obstacles, are transformed from their grid coordinates into world coordinates. These coordinates are stored for navigation purposes and may be visualized using markers for debugging or monitoring.
\end{enumerate}

\begin{figure}[h!]
    \centering
    \includegraphics[width=7.5cm]{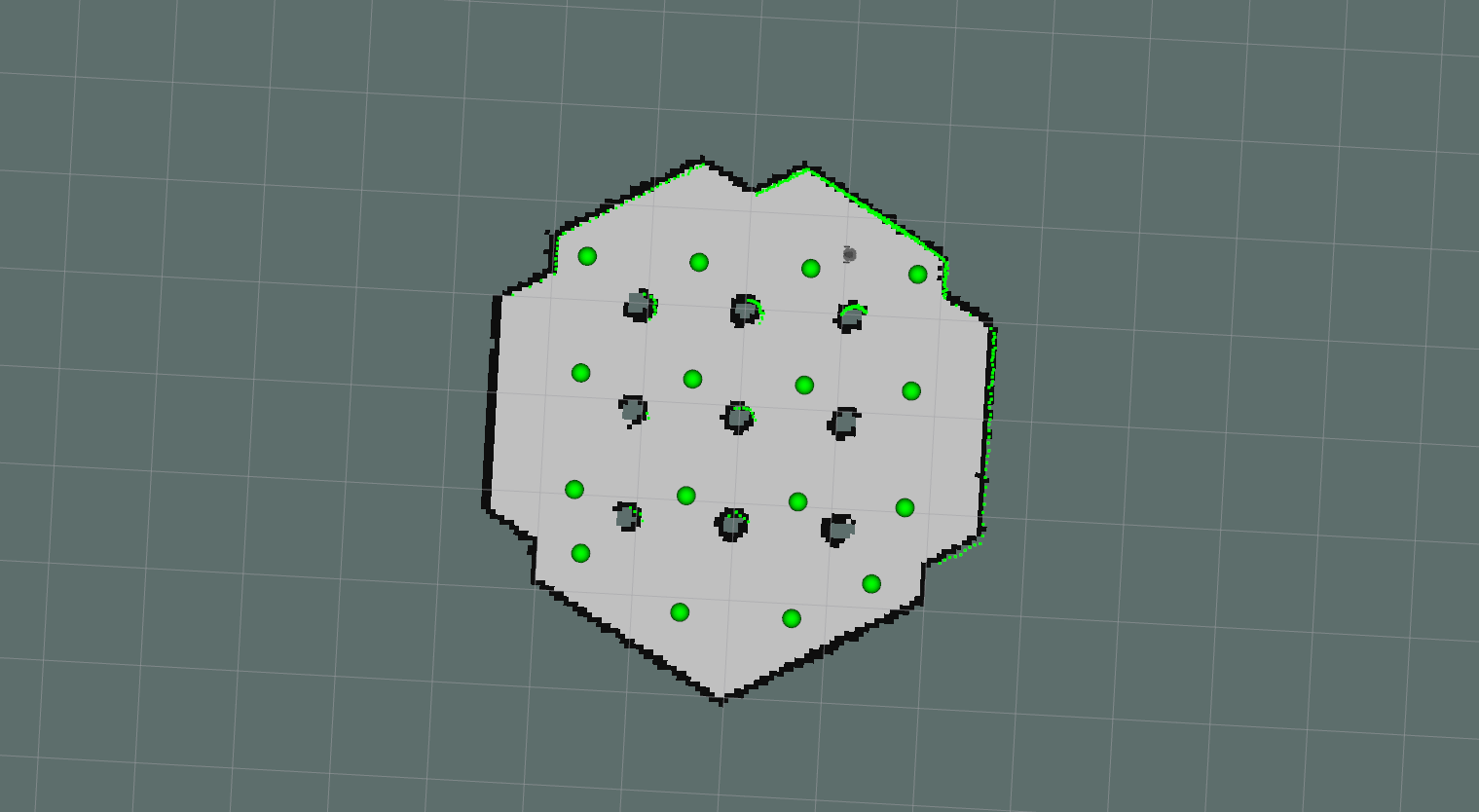}
    \vspace{-0.3cm}
    \caption{Sampled Points in the occupancy grid}
    \label{fig:sampledgridpoints}
\end{figure}

This grid-based path planning approach efficiently explores the free space in the map, ensuring comprehensive coverage while prioritizing robot safety through obstacle avoidance. The generated set of goal points serves as the foundation for the robot's search path, guiding it towards potential locations of AprilTags during the search and rescue operation. The final result of the sampled points in the occupancy is shown in \cref{fig:sampledgridpoints}
\vspace{-0.3cm}

Once all the points are sampled, the node subscribes to the \verb|/tag_detections| topic to continuously receive information about detected AprilTags. Upon receiving data, it parses the information to extract the IDs of the detected AprilTags, storing them in a dictionary. The system continually monitors this dictionary to determine whether all pre-configured AprilTags have been successfully detected.

The sampled points in the queue are sent to the \verb|move_base| action server, responsible for planning a path to the goal and controlling the robot's movements to reach the desired position. Upon reaching each goal point, the robot performs a $360^\circ$ rotation for 4 seconds at the sampled goal point. Once the rotation is complete, the next sampled point is taken from the queue and sent to the \verb|move_base| action server. This process ensures all AprilTags are detected in the environment. Once all AprilTags are detected in the dictionary, the \gls{tb3} ceases exploration.

\section{Results}
\label{sec:results}

\subsection{Survivor Localization}

Due to the challenge of acquiring actual ground truth positions of AprilTags in real-life arenas, we utilize Gazebo to simulate arenas, deliberately placing 11 AprilTags across the map. The ground truth positions of AprilTags can be extracted from the \verb|launch| files of the Gazebo simulated arenas. We specifically examine two simulated arenas: the \gls{tb3} World and the House World.

\subsubsection{TurtleBot3 World}
\begin{figure}[h!]
    \centering
    \includegraphics[width=6.5cm]{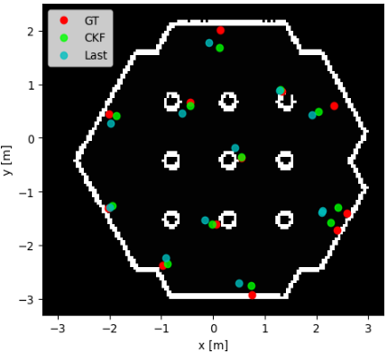}
    \vspace{-0.5cm}
    \caption{Visualize the ground truth positions, based on the TurtleBot3 World SDF launch file, of 11 AprilTags in the Gazebo TurtleBot3 World. Compare these GT positions with the estimated positions from the \gls{ckf} and apriltag\_ros.}
    \label{fig:gazeboworlderror}
\end{figure}

The tb3 World features numerous pillars located in the center of the arena, resulting in occlusion of the AprilTags within the camera's field of view. We hypothesize that this occlusion adversely affects the perspective-n-point transformations used to estimate the relative positions of AprilTags from the raspicam. Consequently, the final measurements from the \verb|april_tag| \gls{ros} node may result in inaccurate position estimates of the AprilTags. This issue is exacerbated as the distance between the AprilTag and the tb3 increases. A visual of the \verb|apriltag_ros| estimator , \gls{ckf} estimator, and ground truth AprilTag position is shown in \cref{fig:gazeboworlderror}

\subsubsection{House}

\begin{figure}[h!]
    \centering
    \includegraphics[width=6.5cm]{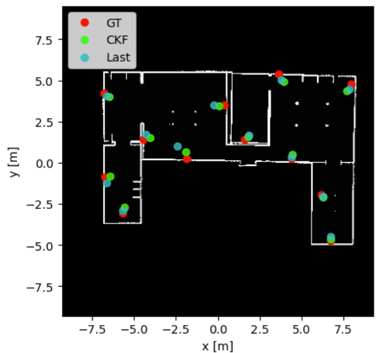}
    \caption{Visualize the ground truth positions, based on the TurtleBot3 House SDF launch file, of 11 AprilTags in the Gazebo TurtleBot3 House. Compare these GT positions with the estimated positions from the \gls{ckf} and apriltag\_ros.}
    \label{fig:gazebohouseerror}
\end{figure}

The House contains open, spacious rooms with minimal obstacles, resulting in little occlusion of the AprilTags within the camera's field of view. However, the absence of occlusion allows the \verb|apriltag_ros| package to detect AprilTags from greater distances, leading to more erroneous measurements compared to the tb3 World arena. We have observed a direct bias estimation error that is proportional to the distance. A visual of the \verb|apriltag_ros| estimator , \gls{ckf} estimator, and ground truth AprilTag position is shown in \cref{fig:gazebohouseerror}

We observed that the \gls{ckf} mitigated most of the issues described above. The \gls{ckf} estimate, which uses many measurements of the same AprilTag, resulted in an improved \gls{mse} for the AprilTag position estimates (\cref{tab:mse_table}). 

\begin{table}[h!]
\centering
  \begin{tabular}{|l|ll|ll|}
    \hline
    \multirow{2}{*}{Tag id} &
      \multicolumn{2}{c|}{World} &
      \multicolumn{2}{c|}{House} \\
    & CKF & Last & CKF & Last  \\
    \hline
    Tag0  & 0.21 & 0.47 & 0.32 & 0.31 \\
    Tag1  & 0.17 & 0.34 & 0.34 & 0.36 \\
    Tag2  & 0.10 & 0.15 & 0.20 & 0.34 \\
    Tag3  & 0.32 & 0.46 & 0.49 & 0.26 \\
    Tag4  & 0.34 & 0.33 & 0.31 & 0.11 \\
    Tag5  & 0.14 & 0.18 & 0.11 & 0.32 \\ 
    Tag6  & 0.10 & 0.03 & 0.22 & 0.27 \\
    Tag7  & 0.10 & 0.24 & 0.34 & 0.37 \\
    Tag8  & 0.05 & 0.24 & 0.28 & 0.42 \\
    Tag9  & 0.07 & 0.26 & 0.28 & 1.2 \\
    Tag10 & 0.19 & 0.46 & 0.37 & 0.18 \\
    Tag11 & 0.06 & 0.04 & 0.33 & 0.23 \\
    \hline
    Avg   & 0.15 & 0.27 & 0.30 & 0.36 \\
    P-Val & \multicolumn{2}{c|}{0.02} & \multicolumn{2}{c|}{0.22} \\
    \hline
  \end{tabular}
  \vspace{0.2cm}
  \caption{Comparing the \gls{mse} (in meters) of the apriltag\_ros package against the \gls{ckf} for the AprilTag position estimate. The last two rows are the average \gls{mse} across all AprilTags, and the unequal variance two-sample Welch t-test.}
  \label{tab:mse_table}
\end{table}

We utilize a two-sample Welch's t-test to determine whether the \gls{ckf} reduction in \gls{mse} is statistically significantly, relative to the \verb|apriltag_ros| package estimator. The Null Hypothesis is the \gls{ckf} \gls{mse} is larger than the last measurement \gls{mse}. For the tb3 World arena, we observe that the \gls{ckf} indeed significantly reduces the \gls{mse} with a P-value of 0.02 (\cref{tab:mse_table}). However, this reduction is not statistically significant for the House area with a P-value of 0.22 (\cref{tab:mse_table}). We hypothesize that with more Monte Carlo trials (and \gls{ckf} hyperparameter optimization), the \gls{ckf} would demonstrate statistical significance in reducing the \gls{mse} for both arenas.

\subsection{Survivors Detected}
We demonstrated that our search and rescue algorithm almost always detects and localizes the AprilTags. However, there were cases where the tb3 failed to detect a single AprilTag (out of 8 AprilTags).
\subsubsection{Gazebo Arenas} In the simulated Gazebo arenas, the tb3 consistently detected and localized all 11 placed AprilTags across three Monte Carlo trials. Examples of successful search and rescue missions in the simulated arena are depicted in Figure \cref{fig:gazeboworlderror} and Figure \cref{fig:gazebohouseerror}. 

\subsubsection{Real World Arena} 
\begin{figure}[htp]
\centering
\subfigure[success]{%
  \includegraphics[clip,width=0.6\columnwidth]{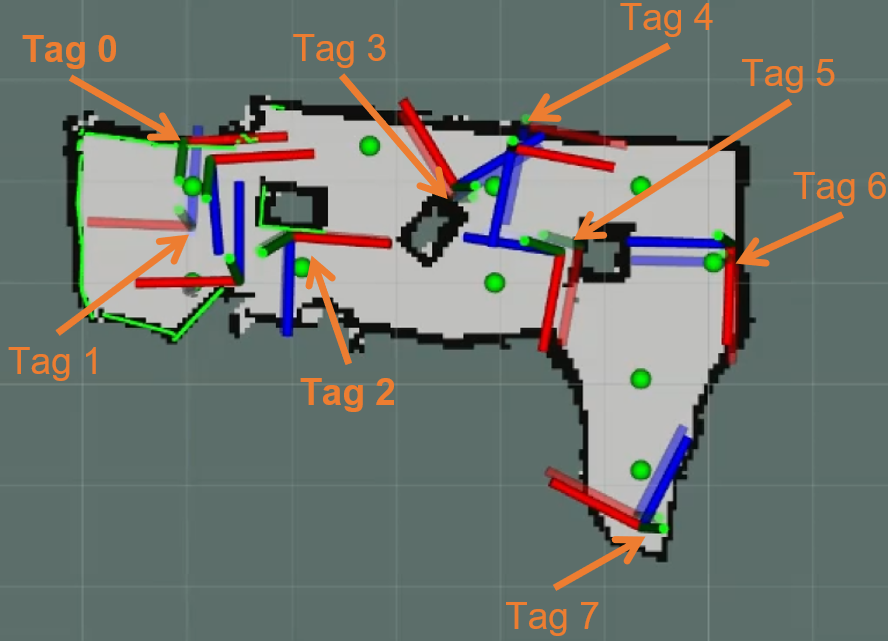}%
}

\subfigure[semi-success]{%
  \includegraphics[clip,width=0.6\columnwidth]{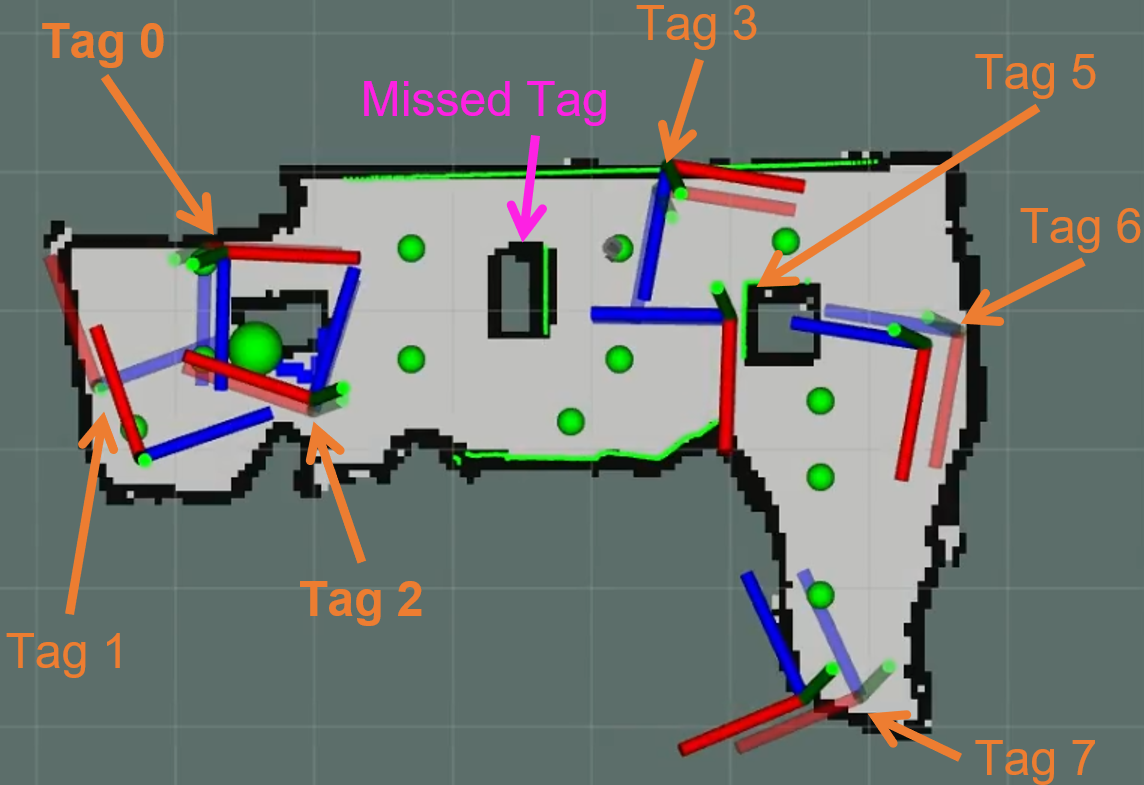}%
}


\caption{(Figure (a) depicts a successful search and rescue mission, where all eight AprilTags are detected and localized. However, Tag1 exhibits poor localization performance, possibly due to the selection of incorrect RVIZ axes. In Figure (b), we observe a semi-successful search and rescue mission, where seven out of eight AprilTags are detected and localized. Upon post-analysis, we discovered that the AprilTag was positioned too high for the camera field of view on the tb3 camera, leading to its failure to detect the AprilTag. The lighter shaded axes are the \gls{ckf} position estimate, whereas the darker axes are the apriltag\_ros last measurement estimate}
\label{fig:apriltag_realworld}
\end{figure}

In our real-world arena (\cref{fig:realworld}), we conducted six search and rescue missions to locate eight scattered AprilTags. Out of the six Monte Carlo trials, two successfully found all eight AprilTags, while four found all but one AprilTag. Upon post-analysis of the unsuccessful Monte Carlo trials, we found that the same AprilTag was consistently not detected. This was attributed to the AprilTag being positioned outside the field of view of the tb3 camera, making it physically impossible for the camera to capture it. 
An example of a successful search and rescue mission, where all eight AprilTags are detected and localized, is illustrated in Figure \cref{fig:apriltag_realworld}(a). Conversely, an unsuccessful search and rescue mission, where only seven out of eight AprilTags were detected and localized, is depicted in Figure \cref{fig:apriltag_realworld}(b).

\subsection{Time to Explore Map}

The \textit{explore\_lite} package and custom exploration package are compared in two gazebo maps: House(\hyperlink{https://youtu.be/_kgmk4pyBCM}{video 1}) and Maze(\hyperlink{https://youtu.be/MxkkJAxtDhY}{video 2}).  The simulated LIDAR has a range 10m and a $360^\circ$ FOV. 
 
The configuration of explore\_lite for both maps is displayed in \cref{tab:explorelite}. The potential scale affects distance to frontier cost, and the gain scale affects frontier size cost. For House gazebo map, it uses the non-inflated map due to exploration failure with costmap. 

\begin{table}[h!]
    \centering
    \begin{tabular}{|c|c|c|}
        \hline
      Parameter & House & Maze \\
        \hline
      map & gmapping map & global costmap \\
      potential scale & 4.0 & 4.0 \\
      gain scale & 1.0 & 1.0 \\
      planner frequency & 0.33 & 0.33 \\
       \hline
    \end{tabular}
    \vspace{0.2cm}
    \caption{Explore Lite Configuration in both maps}
    \label{tab:explorelite}
\end{table}

The configuration (only the most important parameters) of custom exploration (package \verb|exploration_alg|) for both maps is displayed in \cref{tab:customexploration}. The sampling radius represents the square region around each frontier centroid where the algorithm samples points from, and it samples the specified number of samples. The sensor range and field of view affect the computation of information gain and the optimized orientation of goal.  Although the turtlebot LIDAR sensor can scan $360^\circ$, we have limited it so the goal pose always face forward the unknown region.  The move base client requests goal every 3 seconds.

\begin{table}[h!]
    \centering
    \begin{tabular}{|c|c|c|}
        \hline
      Parameter & House & Maze \\
        \hline
      map & global costmap & global costmap \\
      sampling radius [m] & 1.0 & 1.0 \\
      number of samples & 50 & 50 \\
      sensor max range [m] & 7.0 & 7.0 \\
      sensor field of view [$^\circ$] & [-70, 70] & [-70, 70] \\
       \hline
    \end{tabular}
    \vspace{0.2cm}
    \caption{Custom Exploration Configuration in both maps}
    \label{tab:customexploration}
\end{table}

The total exploration time for each map is displayed in \cref{tab:explore}. The custom exploration achieves slightly faster exploration time in both maps.   

\begin{table}[h!]
    \centering
    \begin{tabular}{|c|c|c|}
        \hline
       & explore\_lite & custom exploration \\
        \hline
       House  & 04:44 & 03:05 \\
       \hline
       Maze  & 15:30 & 14:40 \\
       \hline
    \end{tabular}
    \vspace{0.2cm}
    \caption{Exploration time [minutes] comparison}
    \label{tab:explore}
\end{table}

The \textit{explore\_lite} package visualizes frontiers as blue lines and displays its score at the initial cell of each frontier as green spheres. When frontiers are blacklisted due to timeout after a lack of progress, they become red. While it does not directly visualize the chosen goal, its implementation suggests that it selects the frontier centroid.

We observed that \textit{explore\_lite} tends to prioritize large frontiers, even if the \gls{tb3} is far away. Consequently, exploration of the environment often wastes time by traveling to distant regions before completing exploration of local areas. Sometimes, \textit{explore\_lite} selects far frontiers, but due to progress timeout, it abandons these goals and pursues closer frontiers, leading to somewhat erratic behavior that paradoxically improves overall exploration time. Additionally, there are cases where \textit{explore\_lite} fails for unknown reasons.

In contrast, our exploration implementation displays frontiers as blue lines, frontier centroids as blue spheres, best sampled poses near each frontier as yellow arrows, and the chosen exploration goal as a green arrow. Our approach appears to make more reasonable choices in selecting exploration goals, favoring close frontiers with sufficient information gain. However, in certain scenarios, such as in the maze map, our implementation could have explored faster, as evidenced by instances where it chose inefficient farther frontiers for their information gain.

To conclude this section, we highlight the need for considering additional criteria for proper comparison, such as total trajectory length and average information gain. These aspects remain avenues for future work.

\section{Lessons Learned}
Through our experiences, we have gained insights from the challenges encountered. Let us delve into the four notable issues we encountered along the way.
\subsection{Drift Issue}
Drift in mobile robots refers to the unintentional deviation from a planned path or orientation. This can stem from sensor inaccuracies, uneven floor surfaces, or imbalances in the drive mechanism. It's crucial for precision tasks and affects the reliability of autonomous operations. 
When the OpenCR board, housing the IMU, was not securely fastened to the \gls{tb3}, it resulted in odometry readings drifting to infinity, causing the \gls{tb3} to believe it strayed beyond the map boundaries.
\subsection{Carpet Issue}
Operating on carpeted surfaces presents unique challenges for robots, the thickness and texture of carpets can impede movement and operational efficiency. 
We discovered that the wheel encoder inaccurately recorded movement for the \gls{tb3} when the wheels encountered resistance in the carpet, resulting in suboptimal motion control and incorrect odometry.

\subsection{Incorrect Size of April Tag }

\begin{figure}[htp]
\centering
\subfigure[Incorrect AprilTag Edge Size]{%
  \includegraphics[clip,width=0.6\columnwidth]{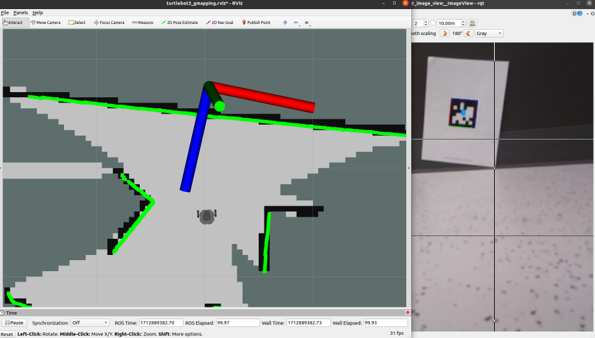}%
}

\subfigure[Correct AprilTag Edge Size]{%
  \includegraphics[clip,width=0.6\columnwidth]{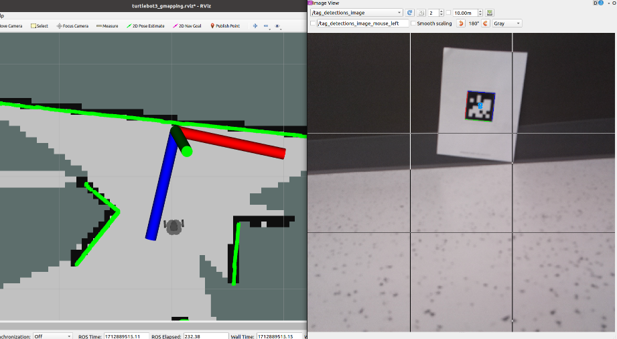}%
}

\label{fig:correctapriltagedgesize}
\end{figure}

We initially set the length of the April Tag to 100 mm, but upon printing, we discovered it was actually 70 mm (refer to \cref{fig:correctapriltagedgesize}(a), where the estimated position is inaccurate). Therefore, we needed to re-measure the actual length to avoid any distortion of the grey box of the QR code during scanning. Figure \cref{fig:correctapriltagedgesize}((b) shows accurate estimation after measuring length of April Tag and camera calibration.

\subsection{Frame Rate Drop}
This issue affects the visual processing capabilities of robots, where the frame rate of camera feeds unexpectedly drops. It can result in delayed image processing and reduced image quality, crucially impacting tasks that require real-time data interpretation.
In real scenario, for the default resolution we desire 15 frames per second. Sometimes it suddenly drops. To maintain the frame rate, we had to reduce resolution and do camera calibration.
\subsection{Hardware Problems}
Robotics systems are susceptible to hardware failures, ranging from motor failures to sensor malfunctions, which can disrupt entire operations. For instance, our OpenCR board experienced a short circuit, necessitating its replacement. Similarly, our LIDAR encountered issues with room scanning, sometimes failing to receive any scans, prompting us to replace it. These steps involved disassembling the \gls{tb3} and reassembling it (and rewiring it) with the new installations.

\section{Future Work}

Our primary focus is on integrating and optimizing our exploration and search and rescue algorithms. Presently, our custom exploration algorithm has only been tested in simulation (Gazebo) and remains disconnected with the search and rescue pipeline. However, we acknowledge the imperative need for these algorithms to collaborate during missions, enabling immediate responses upon detecting victims during exploration. This integration ensures a more efficient and cohesive operation of our robotic system.

Furthermore, we emphasize the importance of extensive testing in both simulated and real-world environments. While our exploration methods have undergone simulation testing with various Gazebo world configurations, real-world scenarios demand further evaluation. By experimenting with different layouts in physical arenas, we can validate the efficacy of our algorithms in practical settings.This holistic approach ensures the robustness and reliability of our system across diverse environments. Additionally, addressing the AprilTag position estimation, the measurement model should either leverage the perspective-n-point model or incorporate the measurement bias error as a corrective function.

\section*{Acknowledgment}

Thanks to Professor Michael Everett for his undying devotion and passion to robotics. Thanks to Donatello for his hardware failing every day.

\bibliographystyle{IEEEtran}
\bibliography{bibliography}

\end{document}